\documentclass{article}
\pdfoutput=1
\PassOptionsToPackage{numbers}{natbib}

\usepackage[final]{nips_2016}

\usepackage[utf8]{inputenc} 
\usepackage[T1]{fontenc}    
\usepackage{url}            
\usepackage{graphicx} 
\usepackage{booktabs}       
\usepackage{amsfonts}       
\usepackage{nicefrac}       
\usepackage{microtype}      
\usepackage{subcaption}
\usepackage{amsfonts}       
\usepackage{amsmath}       
\usepackage{booktabs}
\usepackage{listings}
\usepackage[usenames]{color}
\usepackage{hyperref}       

\definecolor{codegreen}{rgb}{0,0.6,0}
\definecolor{codegray}{rgb}{0.5,0.5,0.5}
\definecolor{codepurple}{rgb}{0.58,0,0.82}
\definecolor{backcolour}{rgb}{0.95,0.95,0.92}

\lstdefinestyle{mystyle}{
    backgroundcolor=\color{backcolour},
    commentstyle=\color{codegreen},
    keywordstyle=\color{magenta},
    numberstyle=\tiny\color{codegray},
    stringstyle=\color{codepurple},
    basicstyle=\footnotesize\ttfamily,
    breakatwhitespace=false,
    breaklines=true,
    captionpos=b,
    keepspaces=true,
    showspaces=false,
    showstringspaces=false,
    showtabs=false,
    tabsize=2
}

\title{Programs as Black-Box Explanations}

\author{
  Sameer Singh\\
  Department of Computer Science\\
  University of California, Irvine CA \\
  \texttt{sameer@uci.edu}
  \And
  Marco Tulio Ribeiro\hspace{15mm} Carlos Guestrin\\
  Department of Computer Science\\
  University of Washington, Seattle WA \\
  \texttt{\{marcotcr, guestrin\}@uw.edu} \\
}

\begin{document}

\maketitle

%

With increasing complexity of machine learning systems being used\footnote{For example, a neural network with a thousand layers was introduced by \citet{he15:deep}.}, there is a crucial need for providing insights into what these models are doing.
\emph{Model-agnostic} approaches~\cite{ribeiro16:model-agnostic}, such as \citet{baehrens10:how-to-explain} and \citet{ribeiro16:kdd}, have shown that insights into complex, black-box models do not have to come at a cost of accuracy, and that accurate \emph{local} explanations can successfully be provided for a number of complex classifiers (such as random forests and deep neural networks) and domains (text and images) for which interpretable models have not performed competitively.
%
However, we still need to identify which interpretable representation would be suitable to convey the local behavior of the model in an accurate and succinct manner, and existing model-agnostic approaches have focused only on (sparse) linear models. 
Work in interpretable machine learning, on the other hand, has proposed many more other representations when designing their models, ranging from additive models, to decision rules, trees, sets, and lists, amongst others~\cite{kim14:the-bayesian,lakkaraju16:interpretable}. 


There are a number of open questions when selecting which of these representations to use for model-agnostic explanations.
It is clear that no single one of these representations, by itself, provides the necessary tradeoff between expressivity and interpretability.
Further, there have not been adequate studies into understanding this tradeoff (with \citet{huysmans11:an-empirical} being an exception), and it is likely that different representations are appropriate for different kinds of users and domains.
It is thus clear that picking any single such intrepretable representation as the choice of model-agnostic representation is not ideal.


In this position paper, we propose using \emph{programs} to explain the local behavior of black-box systems.
There are a number of advantages that such explanations will provide over using any single existing representation.
First, programming languages are designed to capture complex behavior using a high-level syntax that is both succinct and intuitive, and there is a growing group of users that are already trained in reading and writing them.
Second, programs can represent \emph{any} Turing-complete behavior; any of the existing interpretable representation used in literature can be written as a program, but further, programs can also represent arbitrary combinations of multiple of these representations. 
It is also possible to trade off the expressivity and the comprehensibility of the program, for example simple programs for new programmers (at the cost of being an approximation of the complex system) or detailed, longer program for more accurate explanation of the behavior.
Finally, we can potentially apply research in program/software analysis to evaluate various aspects of complex systems, such as automatically characterizing the complexity, security, privacy, and so on.
By providing programs as model-agnostic explanations, we are essentially proposing an approach to \emph{decompile} the local behavior of any black-box, complex systems.

In the following sections, we first demonstrate that program snippets provide a unified yet comprehensible syntax for the commonly used interpretable representations such as decision trees and linear models. 
We then formalize the problem of inducing programs as local explanations of black-box models, and describe a simulated-annealing based prototype implementation.
Finally, we provide examples of generated programs for two datasets using multiple classifiers, demonstrating the expressivity and comprehensibility of using programs to approximate complex, black-box behavior.

\begin{figure}[tb]
    \begin{subfigure}{0.25\textwidth}
        \includegraphics[width=\textwidth]{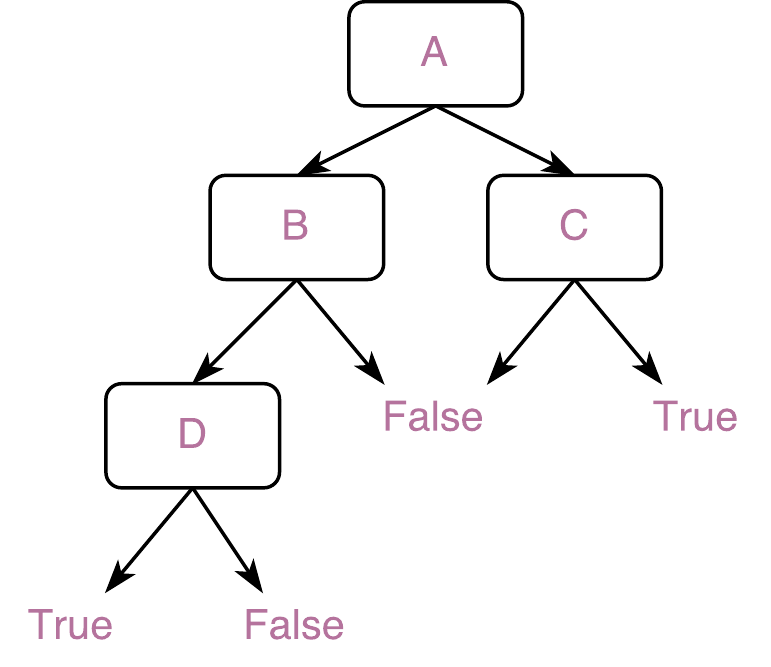}
        \caption{Example decision tree}
        \label{fig:dectree}
    \end{subfigure}
    \qquad
    \begin{subfigure}{0.32\textwidth}
        \begin{lstlisting}[language=Python]


if A:
    if B: return D
    else: return False
else: return not C

        \end{lstlisting}
        \caption{Program for decision tree}
        \label{fig:dectree:prog}
    \end{subfigure}
    \quad
    \begin{subfigure}{0.4\textwidth}
        \begin{lstlisting}[language=Python]

return 10*A - 9*B +2
        \end{lstlisting}
        \begin{lstlisting}[language=Python]

return A or not B
        \end{lstlisting}
        \caption{Equivalent programs for a linear model}
        \label{fig:linprog}
    \end{subfigure}
    \caption{Example decision tree (a) and its corresponding program (b) in our syntax, showing that the complexity is on a similar level. We also show two versions of the linear model, the latter of which is more compact and easier to understand. These programs were manually written.}
    \label{}
\end{figure}

\begin{figure}[b]
    \centering
    \begin{subfigure}{0.75\textwidth}
        \begin{lstlisting}[language=Python]
if RespIllness and Smoker and Age>=50: LungCancer
elif RiskDepression: Depression
elif BMI>=0.2 and Age>=60: Diabetes
elif Headaches and Dizziness: Depression
elif DocVisits>=0.3: Diabetes
elif DispTiredness: Depression
else: Diabetes\end{lstlisting}
        \caption{Decision list}
        \label{fig:declist}
    \end{subfigure}
    \\
    \begin{subfigure}{0.9\textwidth}
        \begin{lstlisting}[language=Python]
if RespIllness and Smoker and Age>=50: LungCancer
if RiskLungCancer and BP>=0.3: LungCancer
if RiskDepression and PastDepression: Depression
if BMI>=0.3 and Insurance=None and BP>=0.2: Depression
if Smoker and BMI>=0.2 and Age>=60: Diabetes
if RiskDiabetes and BMI>=0.4 and ProbInfections>=0.2: Diabetes
if DocVisits>=0.4 and ChildObesity: Diabetes\end{lstlisting}
        \caption{Decision Set}
        \label{fig:decset}
    \end{subfigure}
    \caption{Example of programs for a decision set and a decision list, originally appearing in ~\citet{lakkaraju16:interpretable}, demonstrating that they remain easy to read as programs. These were written manually.}
    \label{fig:decs}
\end{figure}

\section{Other Interpretable Representations as Programs}
\label{sec:prog}

In this section we provide some simple examples of interpretable models currently used in the literature, and describe how their programmatic equivalent would look like.
As it will become apparent, most of these existing interpretable models retain their readability when written as programs.

We use a fairly simple but expressive language for our programs that consists of boolean constants (\texttt{True}, \texttt{False}) and operators (\texttt{and}, \texttt{or}, \texttt{not}), absence/presence of the features of the input instance (\texttt{Smoker}), real valued constants (\texttt{0.5}) and algebraic operators (\texttt{+,-,*}), real-valued features (\texttt{Age}), and if-then-else conditions.
This language is fairly expressive, but due to the lack of looping, recursion, and variables, is still not a complete programming language.
We will use the Python syntax to render our programs, slightly abused to conserve space.

One of the most commonly used interpretable representations is that of decision trees~\cite{craven96:extracting}.
From the simple decision tree in Figure~\ref{fig:dectree}, along with the program for it in Figure~\ref{fig:dectree:prog}, it is clear that the program is a fairly intuitive representation.
Along with decision trees, sparse linear models have also been used in a number of applications as interpretable representations of machine learning~\cite{ustun15:supersparse}.
In Figure~\ref{fig:linprog}, we shows two programs: first that exactly captures the behavior over the relevant features, while the second demonstrates a simpler program that has the same behavior if the features are binary and the prediction is true if the linear model evaluates to a positive score.

Recently, decision lists~\cite{wang15:falling,letham15:interpretable} and decision sets~\cite{lakkaraju16:interpretable} have been introduced as more comprehensible representations than decision trees, while being much more powerful that linear models.
Since these are often presented using pseudo code, the program for these representations in our language looks essentially the same, as shown in Figure~\ref{fig:decs}.


From these examples, it is clear that not only are programs able to represent the different interpretable representations succinctly, but the programming language can be much more expressive than any single representation.
The key challenges is to actually synthesize the appropriate program, i.e. to make sure it is both a good approximation of the black-box model, and is as readable as the examples shown here.
In the next section we will formalize this problem, and describe a prototype solution.

\section{Inducing Program Explanations}
\label{sec:inducing}

In this section, we briefly outline our ideas on how to generate programs as explanations for complex systems, along with the description of a prototype implementation using simulated annealing.

\textbf{Local, Model-Agnostic Explanations:}
Our goal here is to explain individual predictions of a complex machine learning system, by treating them in a black-box manner.
The advantages of generating such model-agnostic explanations was described in \citet{ribeiro16:model-agnostic}.

Our proposed work builds upon the ideas in \citet{ribeiro16:kdd}.
Let the black-box system be $f:\mathcal{X}\rightarrow\{0,1\}$, and we are interested in explaining a specific prediction, i.e. $f(x)=y$.
In order to generate an explanation that describes the behavior of $f$ around $x$, we generate a number of random perturbations of $x$, denoted by $Z$.
We then induce the program that both (1) accurately models the behavior of $f$ on the samples $Z$ (weighed by their similarity to $x$), and (2) is interpretable to the user.
Specifically, we solve the following optimization:
\begin{equation}
    \hat{p} = \underset{p\in\mathcal{P}}{\arg\min}~\mathcal{L}(f, p, Z, \Pi_x) + \Omega(p)\label{eq:opt}
\end{equation}
where $P$ is the set of compatible programs (valid expressions that $\mathcal{X}\rightarrow\{0,1\}$), $\mathcal{L}(f,p,Z,\Pi_x)$ is the loss between the outputs of $f$ and $p$ on the samples $Z$ weighted by $\Pi_x$, and $\Omega(\cdot)$ denotes the complexity of the program (number of lines or the depth of the expression tree, for example).

\textbf{Program Induction:}
Eq~\eqref{eq:opt} is a challenging combinatorial optimization on a potentially complex surface (depending on the loss used).
A related thread of research is \emph{program induction}, where programs are synthesized automatically to match some desired goal~\cite{manna80:a-deductive}.
Number of different variations of this problem have been introduced, depending on the syntax of the program and the formalism of the desired goals, with solvers ranging from genetic programming~\cite{briggs06:functional} to MCMC~\cite{liang10:learning}.
There has also been recent work in using probabilistic programs to identify such programs, mentioned as a possibility in \citet{mansinghka09:natively} using Church~\citep{goodman08:church:}, but with a recent implementation by \citet{gaunt16:terpret:}.
However, we were unable to identify an off-the-shelf program inducer that can support an arbitrary loss (that depends on the domain) in order to identify the appropriate explanation.

\textbf{Prototype Implementation:}
We implemented a prototype program inducer that approximately solves Eq~\eqref{eq:opt} in order to generate program explanations.
We use the same syntax as the one used in Section~\ref{sec:prog}, i.e. boolean constants and operators, input features, real-valued constants and algebraic operators, and if-then-else conditions.
In order to encapsulate the complexity of each program, we set $\Omega(.)$ to be $0$ if the number of nodes in the expression tree is $<$8 and is $\infty$ otherwise, i.e. we are implicitly considering a family of short programs as $\mathcal{P}$.
We use the negative of the weighted $F_1$ score as the loss $\mathcal{L}$, but the implementation supports any arbitrary function that is evaluated on the outputs of $f$ and $p$ on $Z$.
The combinatorial optimization is solved using simulated annealing~\cite{kirkpatrick83:optimization} with a logarithmically decreasing temperature schedule, and a proposal function that randomly grows, shrinks, or replaces nodes in the express tree to create valid perturbed expression trees.

\section{Example Generated Programs}

Using the program induction technique described in the previous section, here we present a few example program explanations for a number of classifiers (treated as black-boxes) on two datasets from the UCI repository~\cite{lichman13:uci-machine}: \emph{adult} and \emph{hospital readmission}.
In order to evaluate whether the programs are accurate as explanations, we also provide a visualization of the decision tree models.

\begin{figure}[tb]
    \begin{subfigure}{0.55\textwidth}
        \includegraphics[width=\textwidth]{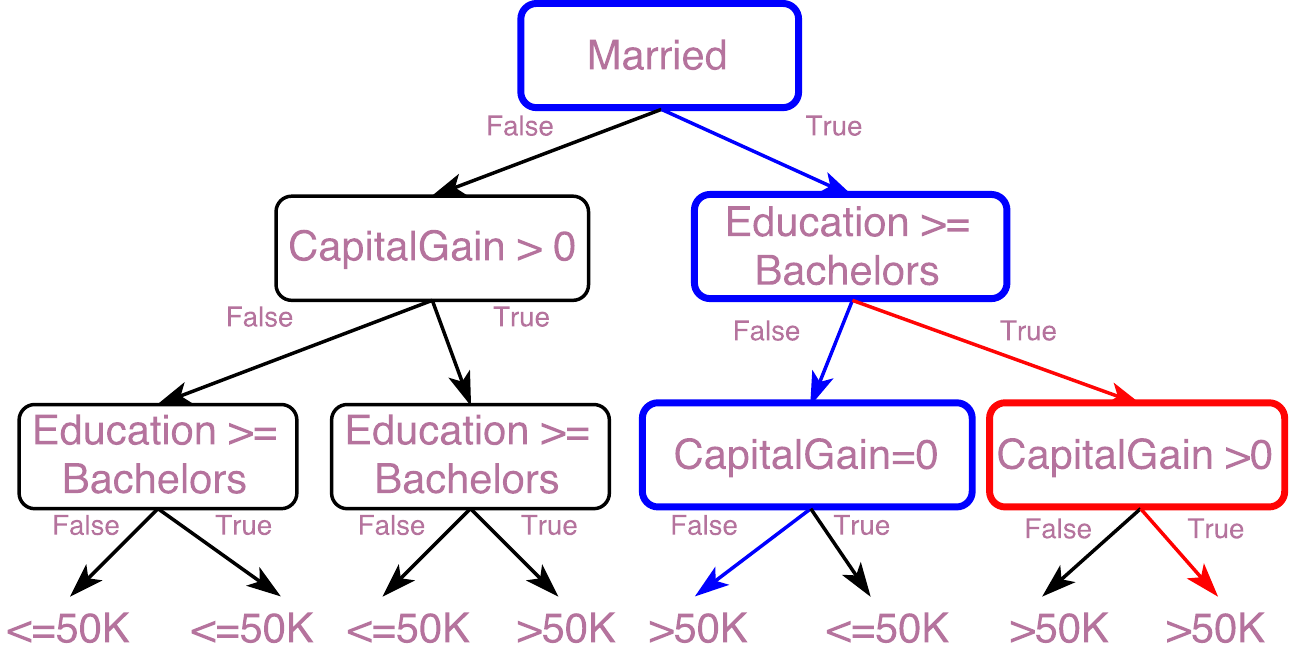}
        \caption{Decision tree (with the path highlighted)}
        \label{fig:adult:tree}
    \end{subfigure}
    \quad
    \begin{subfigure}{0.4\textwidth}
        \textbf{Random Forests} (true implies $>50$K):
        \begin{lstlisting}[language=Python]
(if HoursPerWeek<=40:
   CapitalGain>0
else: True) and Married\end{lstlisting}
        \textbf{Decision Tree} (true implies $>50$K):
        \begin{lstlisting}[language=Python]
CapitalGain>0 and Married\end{lstlisting}
        \textbf{Linear model} (true implies $>50$K):
        \begin{lstlisting}[language=Python]
if CapitalGain>0: Married
else : False\end{lstlisting}
        \caption{Generated program explanations}
        \label{fig:adult:expl}
    \end{subfigure}
    \caption{\textbf{Adult dataset:} In (a), we show the learned tree, with the path for the instance in blue, and in red, we show that \texttt{Education} doesn't really matter for this instance. (b) shows the explanations for three classifiers (they got the prediction right), in particular showing that the explanation for the decision tree gets the more compact form.}
    \label{fig:adult}
\end{figure}

\begin{figure}[tb]
    \begin{subfigure}{0.4\textwidth}
        \includegraphics[width=\textwidth]{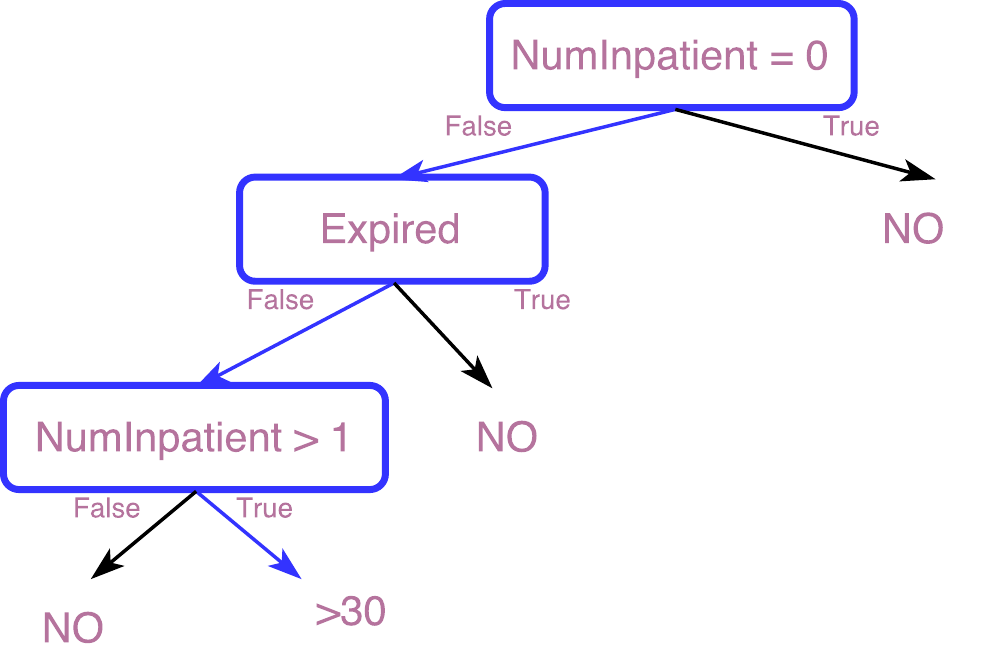}
        \caption{Decision tree (with the path highlighted)}
        \label{fig:readmiss:tree}
    \end{subfigure}
    \qquad
    \begin{subfigure}{0.5\textwidth}
        \textbf{Random Forests} (true implies $<30$):
        \begin{lstlisting}[language=Python]
if Diag:Other and not Tolbutamide:
    Discharged:Home
else: Diag:Other\end{lstlisting}
        \textbf{Decision Tree} (true implies $>30$):
        \begin{lstlisting}[language=Python]
NumInpatient > 1.00\end{lstlisting}
        \textbf{Linear model} (true implies NO):
        \begin{lstlisting}[language=Python]
not Tolazamide\end{lstlisting}
        \caption{Generated program explanations}
        \label{fig:readmiss:expl}
    \end{subfigure}
    \caption{\textbf{Hospital Readmission data:} (a) shows the learned tree, with the path for the instance in blue. Again, (b) shows the explanations for three classifiers (only the tree had the correct prediction), with the compact explanation for tree almost correct, except that it assumes the patient is alive.}
    \label{fig:readmiss}
\end{figure}

In Figures \ref{fig:adult:tree} and \ref{fig:readmiss:tree} we show the learned decision tree on these datasets.
We also trained a random forest classifier and a logistic regression model.
Figures \ref{fig:adult:expl} and \ref{fig:readmiss:expl} show the generated program explanations for both the datasets, demonstrating that the programs are compact and readable, and ones for the decision trees are accurate to the model as well.
Further, it is clear that random forests, which is much more complex in structure than trees or linear models, requires more complicated programs as explanations, however these programs still make sense (Figure \ref{fig:adult:expl}, in particular).

\section{Conclusions and Future Work}

In this paper we motivated the need to use programs as model-agnostic explanations: programs are designed to be intuitive to humans and are incredibly expressive.
We presented a prototype implementation that induces programs as local explanations of a classifier by fitting to the classifier's predictions on a set of perturbations of the instance being explained.
We demonstrated example explanations generated for multiple datasets and classifiers.

There are a number of exciting avenues for future work on these ideas.
We will investigate methods for inducing programs with a much more expressive syntax, including, for example, loops and variables.
Instead of relying on combinatorial optimization techniques that may not scale to applications on more complex domains, syntax, and systems, we will explore the use of recently introduced \emph{differentiable} program induction techniques such as in \citet{neelakantan2015neural} and \citet{riedel16:programming}.
Finally, on real-world applications and using user studies, we will thoroughly evaluate the interpretability and utility of using programs as local explanations of complex machine learning systems.

%


\small
\bibliographystyle{plainnat}
\bibliography{../sameer}

\end{document}